\pdfoutput=1

\documentclass[11pt]{article}

\usepackage{acl}

\usepackage{latexsym}
\usepackage{graphicx}
\usepackage{booktabs}
\usepackage{array}

\usepackage[T1]{fontenc}

\usepackage[utf8]{inputenc}

\usepackage{microtype}

\usepackage{graphicx}
\usepackage{subfigure}
\usepackage{inconsolata}
\usepackage{times}
\usepackage{amsmath}
\usepackage{amssymb}
\usepackage{subfigure}
\usepackage{multirow}

\title{Towards Better Parameter-Efficient Fine-Tuning for Large Language Models: A Position Paper}

\author{Chengyu Wang$^{1}$, Junbing Yan$^{2}$, Wei Zhang$^{2}$, Jun Huang$^{1}$\\
  $^{1}$ Alibaba Group, Hangzhou, China \\
  $^{2}$ East China Normal University, Shanghai, China \\
  \texttt{\{chengyu.wcy,huangjun.hj\}@alibaba-inc.com,}\\
  \texttt{\{junbingyan531,zhangwei.thu2011\}@gmail.com}}

\begin{document}
\maketitle
\begin{abstract}
This paper delves into the pressing need in Parameter-Efficient Fine-Tuning (PEFT) for Large Language Models (LLMs). While LLMs possess remarkable capabilities, their extensive parameter requirements and associated computational demands hinder their practicality and scalability for real-world applications.
Our position paper highlights current states and the necessity of further studying into the topic,
and recognizes significant challenges and open issues that must be addressed to fully harness the powerful abilities of LLMs.
These challenges encompass novel efficient PEFT architectures, PEFT for different learning  settings, PEFT combined with model compression techniques, and the exploration of PEFT for multi-modal LLMs. By presenting this position paper, we aim to stimulate further research and foster discussions surrounding more efficient and accessible PEFT for LLMs.
\end{abstract}

\section{Introduction}

Large Language Models (LLMs) have exhibited remarkable capabilities, with popular models such as ChatGPT\footnote{\url{https://openai.com/blog/chatgpt}} and GPT4~\cite{DBLP:journals/corr/abs-2303-08774} showcasing their potentials in a variety of NLP tasks~\cite{DBLP:journals/corr/abs-2303-18223,DBLP:journals/corr/abs-2307-04251}. 
However, these LLMs often suffer from  extensive parameter requirements and associated computational demands, limiting their practicality and scalability for real-world applications. Parameter-Efficient Fine-Tuning (PEFT) addresses the challenges by reducing the number of parameters required for effective fine-tuning without compromising the model performance.
Notable PEFT approaches include LoRA~\cite{DBLP:conf/iclr/HuSWALWWC22}, adapter tuning~\cite{DBLP:conf/icml/HoulsbyGJMLGAG19}, prefix-tuning~\cite{DBLP:conf/acl/LiL20}, prompt-tuning~\cite{DBLP:conf/emnlp/LesterAC21}, P-tuning~\cite{DBLP:conf/acl/LiuJFTDY022}, BitFit~\cite{DBLP:conf/acl/ZakenGR22} and others.

Despite the advancement, we observe that existing PEFT approaches for LLMs present several limitations that hinder their effectiveness and practicality. Most PEFT methods are proposed in the BERT era primarily for encoder-based models, which are not tailored specifically for LLMs. Detailed PEFT implementations are mostly agnostic without considering the decoder-only architectures and the algorithmic characteristics of mainstream LLMs, for example, the requirements of Reinforcement Learning from Human Feedback (RLHF)-based fine-tuning~\cite{DBLP:conf/nips/Ouyang0JAWMZASR22}.
Thus, there is an urgent need for better PEFT that enables more effective learning of LLMs.

In this position paper, we advocate for the development of PEFT techniques specifically tailored for LLMs. We briefly review current states of development in the field. Based on our empirical study, we show that in general LoRA-based approaches are more suitable for LLMs; yet there are no uniform algorithmic designs for all the settings. In addition, we discuss complicated learning strategies that are not supported by current PEFT methods, such as more efficient distributed PEFT, PEFT that support RLHF training for better human alignment, PEFT combines with various model compression techniques (such as distillation and quantization), and PEFT for multi-modal LLMs. We hope that our research can stimulate research for better PEFT techniques, especially for LLMs.

\section{Literature Review}


\noindent\textbf{A Brief Overview of LLMs.}
Before the LLM wave, Pre-trained Language Models (PLMs) have gained significant attention due to their abilities to learn contextual representations~\cite{DBLP:journals/corr/abs-2003-08271,DBLP:journals/corr/abs-2111-01243}. 
One prominent example is BERT~\cite{DBLP:conf/naacl/DevlinCLT19}, which leverages the encoder-only architecture and has been adopted in language understanding tasks.
Since the launch of ChatGPT, a variety of LLMs have been released. Popular open LLMs include LLaMA~\cite{DBLP:journals/corr/abs-2302-13971}, LLaMA 2~\cite{DBLP:journals/corr/abs-2307-09288}, OPT~\cite{DBLP:journals/corr/abs-2205-01068}, 
OPT-IML~\cite{DBLP:journals/corr/abs-2212-12017},
GPT-NeoX~\cite{DBLP:journals/corr/abs-2204-06745}, BLOOM~\cite{DBLP:journals/corr/abs-2211-05100},
BLOOMZ~\cite{DBLP:conf/acl/MuennighoffWSRB23},
Galactica~\cite{DBLP:journals/corr/abs-2211-09085}, CPM-2~\cite{DBLP:journals/aiopen/ZhangGHCXSYQGKC21}, GLM~\cite{DBLP:conf/iclr/ZengLDWL0YXZXTM23}, Pythia~\cite{DBLP:conf/icml/BidermanSABOHKP23}, and many others, to name a few. 
For model training, the three stage process of ``pre-training, supervised fine-tuning (SFT) and RLHF'' put forward by~\cite{DBLP:conf/nips/Ouyang0JAWMZASR22} is widely accepted by the community. 
It can be easily seen that training LLMs requires numerous computational resources. Therefore, the huge computational and financial costs naturally call for the development of PEFT for LLMs.

\noindent\textbf{General PEFT Methods.}
PEFT is a type of fine-tuning methods that reduce
the number of learnable parameters of PLMs (not specifically for LLMs) while preserving good
performance, which is also referred to as Delta Tuning~\cite{DBLP:journals/natmi/DingQYWYSHCCCYZWLZCLTLS23}. 
BitFit~\cite{DBLP:conf/acl/ZakenGR22} is a simple sparse fine-tuning method where  only the bias parameters are tuned.
LoRA~\cite{DBLP:conf/iclr/HuSWALWWC22} leverages low-rank approximation to  the update matrices (i.e., parameters) at each model layer, which can be applied to various PLMs. Following the work of LoRA, AdaLoRA~\cite{DBLP:conf/iclr/ZhangCBH0CZ23} is proposed to incorporate adaptive budget allocation into the choices of LoRA ranks for different matrices. DyLoRA~\cite{DBLP:conf/eacl/ValipourRKG23} further employs a dynamic search-free technique for rank selection.
Adapters~\cite{DBLP:conf/icml/HoulsbyGJMLGAG19} are small neural network modules integrated into original transformer blocks, which are learned to capture new knowledge for downstream tasks.
AdaMix~\cite{DBLP:conf/emnlp/WangAM00AG22} learns a mixture of multiple adapters for PEFT.
Prefix-tuning~\cite{DBLP:conf/acl/LiL20} adds a sequence of prefixes represented as trainable continuous embeddings to each transformer layer that specifically capture the task-specific information.
Adaptive Prefix-tuning~\cite{DBLP:conf/acl/ZhangTX00H23} extends Prefix-tuning to make the lengths of prefixes more adaptive to tasks.
P-tuning v2~\cite{DBLP:conf/acl/LiuJFTDY022} is a similar approach that shows layer-wise prompt vectors are also beneficial for solving language understanding tasks. Prefix Propagation~\cite{DBLP:conf/acl/0006ABZ23} explores prefix-tuning for longer input sequences.
In contrast to continuous vectors, prompt-tuning employs trainable
prompt vectors~\cite{DBLP:conf/emnlp/LesterAC21,DBLP:journals/corr/abs-2103-10385,DBLP:conf/emnlp/0001WQH021,DBLP:conf/wsdm/Xu0QLXHH23} or discrete textual descriptions~\cite{DBLP:conf/emnlp/ShinRLWS20,DBLP:conf/acl/GaoFC20} at the input layer to model task-level knowledge. We refer reader to the survey~\cite{DBLP:journals/csur/LiuYFJHN23} for a more detailed review.

\noindent\textbf{PEFT Methods for LLMs.}
It is worth noting that the above methods are not tailored to LLMs. Thus, we further summarize how these PEFT techniques are applied.
To the best of our knowledge, LoRA~\cite{DBLP:conf/iclr/HuSWALWWC22} is one of the most widely applied methods due to its simplicity in design and uniformity in application scenarios. 
Apart from LoRA, LLaMA-Adapter~\cite{DBLP:journals/corr/abs-2303-16199} is proposed to insert adapter networks into LLMs with zero-initialized attention.
In the open-source community, PEFT\footnote{\url{https://github.com/huggingface/peft}} is also the name of a project that provides the implementations of several PEFT methods on LLMs, serving as a useful tool for further research into the subject. 
OpenDelta~\cite{DBLP:conf/acl/HuDZLZLS23} focuses on the quick adaptation of LLMs.
A few works focus on evaluating PEFT on text generation tasks~\cite{DBLP:conf/emnlp/Chen0ML22,DBLP:conf/emnlp/0008PPAPPLSC22}, but are not conducted for LLMs.
Another thread of works combine model quantization with PEFT, which maps model parameters from floating-point numbers to integers~\cite{DBLP:journals/corr/abs-2103-13630}.
QLoRA~\cite{DBLP:journals/corr/abs-2305-14314} quantizes an LLM to 4-bit, and then leverages a small set of LoRA weights to avoid performance degradation.
Alpha Tuning~\cite{DBLP:conf/emnlp/KwonKBYKPK0SL22} and QA-LoRA~\cite{DBLP:journals/corr/abs-2309-14717} are quantization-aware adaptation methods for LLMs.
AWQ~\cite{DBLP:journals/corr/abs-2306-00978} significantly reduces the model quantization error by protecting 1\% of the salient weights of the LLM.

\begin{table}
\centering
\begin{footnotesize}
\begin{tabular}{l | llll } 
\hline
\bf Metric  & \bf FT & \bf LoRA & \bf Prompt & \bf Prefix\\
\hline
\multicolumn{5}{l}{\it Dataset: E2E (Text Generation)}\\
\hline
BLEU-1 & 0.5460 & 0.5000 &0.4476 &0.4552\\
BLEU-2 & 0.3956 &0.3486&0.2994&0.3252\\
METEOR & 0.3448&0.3265&0.2816&0.2952\\
ROUGE-L & 0.3918&0.3569&0.3153&0.3312\\
CIDEr & 0.9502&0.7646&0.5163&0.6003\\
\hline
\multicolumn{5}{l}{\it Dataset: WebNLG (Text Generation)}\\
\hline
BLEU-1 & 0.3025&0.3217&0.2352&0.2587\\
BLEU-2 & 0.2109&0.2173&0.1943&0.2005\\
METEOR & 0.2014&0.1992&0.1698&0.1754\\
ROUGE-L & 0.3045&0.2881&0.2398&0.2465\\
CIDEr & 0.6207&0.5029&0.3465&0.4186\\
\hline
\multicolumn{5}{l}{\it Dataset: GSM8K (Math Problem)}\\
\hline
Accuracy & 0.2382&0.21542&0.15643&0.17454\\
\hline
\multicolumn{5}{l}{\it Dataset: CoQA (Question Answering)}\\
\hline
EM & 0.6089&0.5976&0.5193&0.5339\\
F1 & 0.7004&0.6958&0.5930&0.6264\\
\hline
\end{tabular}
\end{footnotesize}
\caption{Performance of PEFT methods and FT over multiple generation tasks. Note: FT (full fine-tuning), Prompt (prompt-tuning), Prefix (prefix-tuning).}
\vspace{-.5em}
\label{tab:model}
\end{table}

\section{Analysis and Research Directions}

We analyze the performance of PEFT on LLMs and suggest several directions for future research.

\subsection{Empirical Analysis}

Before presenting research directions, we conduct a brief empirical analysis on the effectiveness of PEFT over LLMs. Without loss of generality, we evaluate the performance of a popular LLM, i.e., Llama-2-7b-chat\footnote{\url{https://huggingface.co/meta-llama/Llama-2-7b-chat-hf}}, over two text generation tasks (E2E~\cite{DBLP:conf/sigdial/NovikovaDR17} and WebNLG~\cite{DBLP:conf/acl/GardentSNP17} and two more challenging task of math problems (GSM8K~\cite{DBLP:journals/corr/abs-2110-14168} and question answering (CoQA~\cite{DBLP:journals/tacl/ReddyCM19}). Detailed dataset statistics and experimental settings can be found in the appendix. 

In Table~\ref{tab:model}, we report the testing performance of standard fine-tuning and three popular PEFT methods, namely, prompt-tuning~\cite{DBLP:conf/emnlp/LesterAC21}, prefix-tuning~\cite{DBLP:conf/acl/LiL20} and LoRA~\cite{DBLP:conf/iclr/HuSWALWWC22}. Results show that 
LoRA and full fine-tuning exhibit similar performance in generative tasks, with minimal differences in quality of generated contents. For math problems and QA tasks, they display a slight variance in accuracy, whereas other PEFT methods perform inadequately.
In Table~\ref{tab:larger_models} we study the effectiveness of LoRA on a larger model scale based on Llama-2 and Vicuna\footnote{\url{https://lmsys.org/blog/2023-03-30-vicuna/}}. The results indicate that for simple text generation, there is a minor enhancement as the model size increases from 7B to 13B. However, there is no discernible difference in the generation quality of specific questions after manual checking. Conversely, for more intricate math problems, we observe a significant improvement in accuracy with the increase in model parameters. 

\begin{figure}
\centering
\includegraphics[width=0.55\linewidth]{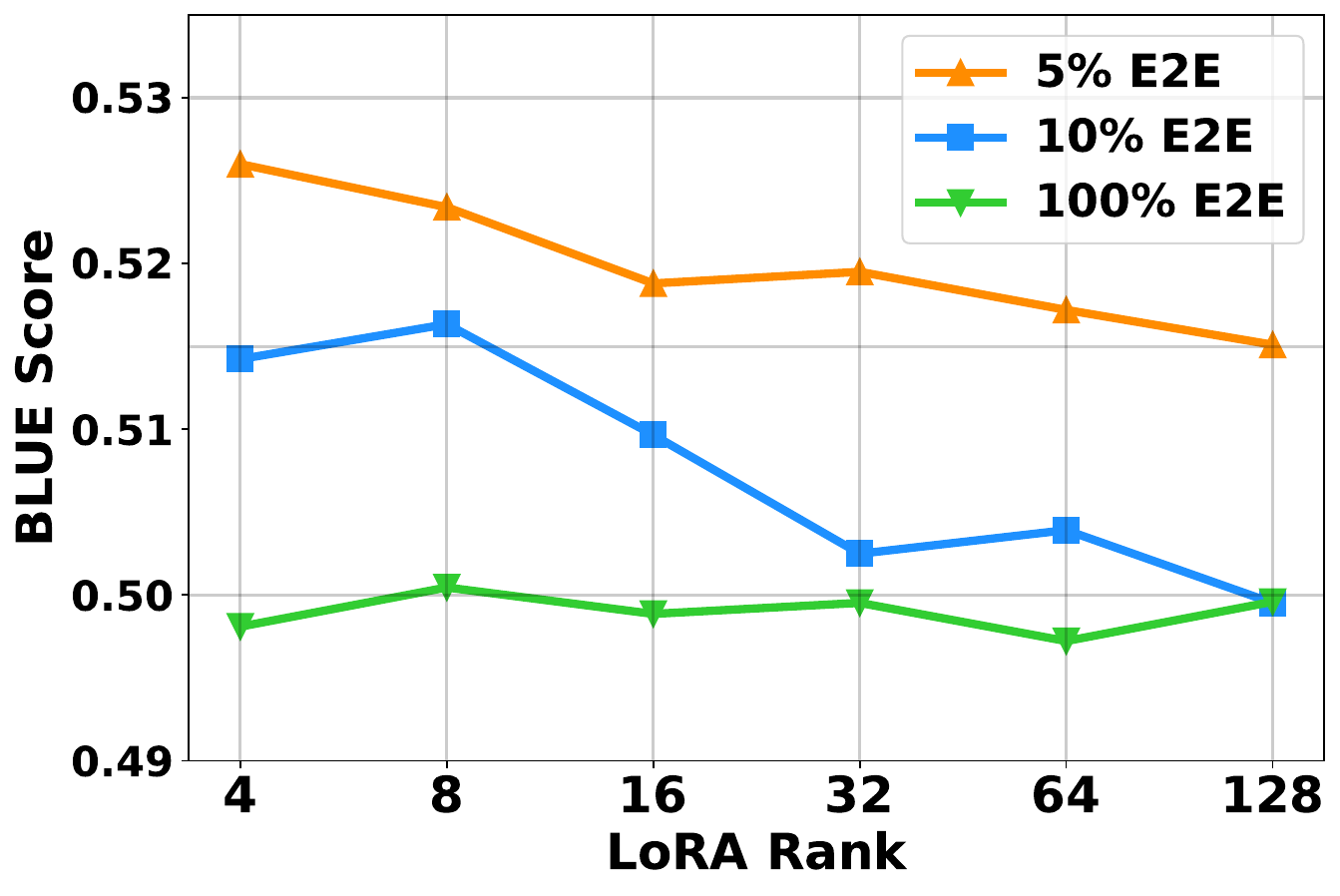}
\caption{The impact of data volume  ($5\%$, $10\%$, $100\%$ of the E2E dataset) with different LoRA ranks.}
\vspace{-.75em}
\label{fig:lora_rank}
\end{figure}

We further observe that different LoRA ranks have varying degrees of performance impact. To investigate this further, we conduct tests on the same dataset with different data volumes (randomly sampled 5\%, 10\%, and the entire dataset) using different LoRA ranks. As shown in Figure~\ref{fig:lora_rank}, for smaller datasets, a lower LoRA rank yields optimal results, and increasing LoRA ranks actually leads to a decline in performance. 
Therefore, a lower LoRA rank can achieve satisfactory performance while also saving training resource costs.


\begin{table}
\centering
\begin{footnotesize}
\begin{tabular}{l | llll } 
\hline
\bf Metric  & \bf Llama-2 & \bf Llama-2 & \bf Vicuna & \bf Vicuna\\
& \bf (7B) & \bf (13B) & \bf (7B) & \bf (13B)\\
\hline
\multicolumn{5}{l}{\it Dataset: E2E (Text Generation)}\\
\hline
BLEU-1 & 0.5000 &0.5028&0.5038&0.5066\\
BLEU-2 &0.3486&0.3522&0.3526&0.3545\\
METEOR &0.3265&0.3228&0.3238&0.3247\\
ROUGE-L &0.3569&0.3559&0.3558&0.3560\\
CIDEr &0.7646&0.7986&0.7783&0.8106\\
\hline
\multicolumn{5}{l}{\it Dataset: GSM8K (Math Problem)}\\
\hline
Accuracy & 0.2382&0.3835&0.1641&0.2273\\
\hline
\end{tabular}
\end{footnotesize}
\caption{LoRA performance with different model sizes.}
\vspace{-.5em}
\label{tab:larger_models}
\end{table}

\subsection{Lessons Learned for Future Research}

From the experiments, it is seen that LoRA-style PEFT methods achieve better performance for LLMs. Yet, there is no ``free lunch'' for all learning settings, particularly for different tasks and data volumes. In addition, the trained LoRA modules with large ranks may still be over-parameterized for some cases. We suggest that for future research, here are some possible directions.
i) Task-adaptive LoRA methods can be developed to search more suitable ranks based on task difficulty and data volumes.
ii) More compact low-rank structures can be involved to decompose the parameter matrices, which speed up the training process and avoid over-fitting simultaneously.
iii) Combining LoRA-style approaches with better prompt designs for LLMs may also result in better performance.

\subsection{Other Research Directions}

\noindent\textbf{Large-scale Training.}
As observed from the experimental results, the performance of LoRA is highly related to the number of trainable parameters (controlled by the LoRA rank), for LLMs with 100B parameters or more (such as GPT-4~\cite{DBLP:journals/corr/abs-2303-08774}), even turning only 1\% of the parameters leads to huge computational costs. In addition, the model checkpoints must be partitioned as they do not fit in single GPU. Thus, the parameters LoRA of modules are also distributed according to the model partition strategies during training. The parameter values should be communicated frequently across GPUs and machines during the training process. To the best of our knowledge, there are no comprehensive studies or publicly available frameworks that address the problems of large-scale, distributed LoRA training for ultra-large models effectively.

In addition, the auto-regressive language model next token prediction objective is not the only learning task during the LLM training process. For better alignment with human values, RLHF~\cite{DBLP:conf/nips/Ouyang0JAWMZASR22} is often leveraged to fine-tune the LLMs based on reinforcement learning coached by a reward model. This process is more computationally expensive due to the involvement of both the supervised fine-tuned and RLHF-based fine-tuned LLM checkpoints, together with a reward model that expresses human preferences. Compared to simple fine-tuning, RLHF requires the computational graphs and weights of these additional models loaded into the GPU memory during training, which significantly lowers the GPU memory space for training the LLM itself. We suggest that further studies on PEFT-style RLHF training is of greater value to save computational resources and benefit the NLP community for deeper research into how to apply RLHF more easily.

\noindent\textbf{PEFT with Model Compression.}
For application developers, it is more important to deploy LLMs online for real-time inference. Hence, compressing LLMs to smaller sizes is critical, in order to save GPU memory and speedup the inference process. 
In the literature, several types of approaches have been proposed to compress the models, such as knowledge distillation, model quantization and pruning.
Take quantization as example. In QLoRA~\cite{DBLP:journals/corr/abs-2305-14314}, the underlying LLM is quantized to 4-bit first and then tuned using LoRA over a small but high-quality dataset.
The work~\cite{DBLP:conf/acl/HsiehLYNFRKLP23} distills LLMs by extracting rationales as additional supervision from larger models for training small models, yet the parameters of small models require to be fully fine-tuned to ensure high performance. 
LLM-Pruner~\cite{DBLP:journals/corr/abs-2305-11627} leverages structural pruning for LLMs to selectively removes non-critical structures based on the gradients learned during training.
A similar work Wanda~\cite{DBLP:journals/corr/abs-2306-11695} prunes weights smallest magnitudes multiplied by the corresponding input activations, in order to bring parameter sparsity to large models.
We suggest that the research on LLM compression with PEFT is vital for online deployment and highly insufficient in current states.
For example, it is possible to obtain a smaller model by PEFT-applied distillation. This benefits institutions or developers where fully fine-tuning smaller models (with parameters around 7B) is computationally prohibitive.

\noindent\textbf{PEFT for Multi-modal LLMs.}
LLMs are not only about texts. By feeding the output representations of visual encoders (or encoders for other modalities) into LLM backbones, multi-modal LLMs, including NExt-GPT~\cite{DBLP:journals/corr/abs-2309-05519}, InstructBLIP~\cite{DBLP:journals/corr/abs-2305-06500}, mPLUG-Owl~\cite{DBLP:journals/corr/abs-2304-14178}, LLaVa~\cite{DBLP:journals/corr/abs-2304-08485}, MiniGPT~\cite{DBLP:journals/corr/abs-2310-09478} and many others, can be trained and deployed to tackle multi-modal tasks by instruction following. In multi-modal LLMs, unifying the representations of different modalities into the same semantic space is crucial for multi-modal understanding and generation. For instance, InstructBLIP~\cite{DBLP:journals/corr/abs-2305-06500} leverages a Q-Former to extract instruction-aware visual features as the input to a frozen LLM. However, without the training of the LLM, it obtains no new knowledge on how to solve the multi-modal tasks.
We believe that PEFT can act as the ``bridge'' to achieve cross-modal communications by slightly tuning existing LLMs that effectively prevents the catastrophic forgetting of uni-modal knowledge.

\noindent\textbf{Other Topics.}
In addition to the above mentioned topics, there are other topics that are worth exploring. Strategies such as adaptive learning rates and regularization methods specifically designed for PEFT can further faster and stabilize the training process. Apart from RLHF, examining parameter-efficient ways to address ethical considerations, such as knowledge securities, fairness or bias mitigation~\cite{DBLP:journals/corr/abs-2307-16680}, can contribute to the development of more reliable and unbiased LLMs. Due to space limitation, we do not elaborate.

\section{Concluding Remarks}

In this position paper, we have highlighted the pressing need for better PEFT methods tailored for LLMs, underscoring the importance of addressing the challenges and open issues in PEFT and encompassing the exploration of novel efficient PEFT architectures, PEFT for different learning settings, and PEFT for multi-modal LLMs. By addressing these challenges, we can pave the way for more efficient and accessible PEFT techniques that are more practical for real-world applications.

\newpage

\section*{Limitations}

This paper is a position paper and does not present any specific new methodologies or approaches  that could be employed to tackle the identified challenges. The limitations and research directions mentioned in the paper are based on the authors' perspectives and may not encompass the entire scope of issues related to PEFT for LLMs.

\appendix

\section{Datasets and Experimental Settings}

\noindent\textbf{Datasets.} 
We evaluate the results on two standard neural generation datasets for the table-to-text task: E2E~\cite{DBLP:conf/sigdial/NovikovaDR17}, WebNLG~\cite{DBLP:conf/acl/GardentSNP17}, one math problem reasoning dataset: GSM8K~\cite{DBLP:journals/corr/abs-2110-14168} and one QA dataset CoQA~\cite{DBLP:journals/tacl/ReddyCM19}. 

Specifically, the E2E dataset contains approximately 50K examples featuring 8 distinct fields. It includes multiple test references for each source table and has an average output length of 22.9. We employ the official evaluation script, which provides metrics such as BLEU, METEOR, ROUGE-L and CIDEr for assessment.
The WebNLG dataset consists of 22K examples where the input $x$ consists of sequences of (subject, property, object) triples. The average output length is 22.5. The training and validation splits encompass input descriptions of entities from 9 distinct DBpedia categories, such as Monuments.
The test split is divided into two sections: the first half contains categories observed in the training data, while the second half includes 5 unseen categories for extrapolation evaluation. For evaluation, we also utilize the official evaluation script.
GSM8K presents a challenging arithmetic reasoning task that language models frequently find difficult to tackle.
CoQA is a challenging task to measure the model abilities to understand a text passage and answer a series of related questions.

\noindent\textbf{Experimental Settings.}
The experiments are conducted on a Linux server with two NVIDIA A100-80G GPUs. We choose Llama-2-7b-chat as the default LLM. In addition, Llama-2-13b-chat, together with the 7B and 13B versions of the Vicuna models, is leveraged for study.

\noindent\textbf{Hyper-parameter Settings.} At training time, we use AdamW \citep{AdamW} as the optimizer, and set its hyper-parameter $(\beta1, \beta2)$ to (0.9, 0.98). The hyper-parameters we tune
include the number of epochs, the batch size, the learning rate, and the sequence length. Hyper-parameter details are in shown in Table~\ref{tab:diff hyper} and Table~\ref{tab:same hyper}. 

\begin{table}
\centering
\begin{small}
\begin{tabular}{l | ccc} 
\hline
\textbf{Dataset} & \bf Epoch  & \bf Sequence Length\\ 
\hline
E2E & 10 & 256 \\ 
WebNLG & 10  & 256\\
GSM8K & 10  & 512\\ 
CoQA & 5  & 2048\\
\hline
\end{tabular}
\end{small}
\caption{Hyper-parameter settings for individual datasets.} 
\label{tab:diff hyper}
\end{table}

\begin{table}[!t]
\centering
\begin{small}
\begin{tabular}{lc}
\hline
& \bf Value  \\ \hline 
Learning Rate & 3e-6 \\
AdamW ($\beta_1$, $\beta_2$) &  (0.9, 0.98) \\ 
Dropout & 0.1 \\ 
Weight Decay & 0.01 \\
Batch Size & 48 \\
\hline
\end{tabular}
\end{small}
\caption{Hyper-parameter settings for all datasets.}
\label{tab:same hyper}
\end{table}

\end{document}